\documentclass{article}
\usepackage[final]{neurips_2021}
\usepackage[hyphens]{url}
\usepackage{graphicx}
\graphicspath{ {./images/} }
\usepackage{xcolor}
 
\usepackage{placeins}
\usepackage{amsfonts}

\usepackage{amsmath}
\usepackage{booktabs} 
\usepackage{arydshln} 

\usepackage{listings}

\usepackage{pifont}
\newcommand{\cmark}{\text{\ding{51}}}
\newcommand{\xmark}{\text{\ding{55}}}

\usepackage{hyperref}

\title{Counter-Strike Deathmatch \\ with Large-Scale Behavioural Cloning}

\author{%
  Tim Pearce$^{1,2}$\thanks{Project started while affiliated with University of Cambridge, now based in Tsinghua University.}  $\;\;\;$ Jun Zhu$^{1}$ \\
   $^1$Tsinghua University $^2$University of Cambridge \\
}

\begin{document}

\maketitle

\begin{abstract}

This paper describes an AI agent that plays the modern first-person-shooter (FPS) video game `Counter-Strike; Global Offensive' (CSGO) from pixel input. The agent, a deep neural network, matches the performance of the medium difficulty built-in AI on the deathmatch game mode whilst adopting a humanlike play style. Previous research has mostly focused on games with convenient APIs and low-resolution graphics, allowing them to be run cheaply at scale. This is not the case for CSGO, with system requirements 100$\times$ that of previously studied FPS games. This limits the quantity of on-policy data that can be generated, precluding many reinforcement learning algorithms. Our solution uses behavioural cloning — training on a large noisy dataset scraped from human play on public servers (5.5 million frames or 95 hours), and smaller datasets of clean expert demonstrations. This scale is an order of magnitude larger than prior work on imitation learning in FPS games. To introduce this challenging environment to the AI community, we open source code and datasets.

\begin{center}
\textbf{Four minute introduction:}
\textcolor{blue}{\url{https://youtu.be/rnz3lmfSHv0}}  \\
\textbf{Gameplay examples:}
\textcolor{blue}{\url{https://youtu.be/KTY7UhjIMm4}} \\
\textbf{Code, model \& datasets:}
\textcolor{blue}{\url{https://github.com/TeaPearce}}
\end{center}

\end{abstract}


\section{Introduction}
\label{sec_intro}

Deep neural networks have achieved strong performance in a variety of video games; from 1970's Atari classics, to 1990's first-person-shooter (FPS) titles Doom and Quake III, and modern real-time-strategy games Dota 2 and Starcraft II \citep{Mnih2015a, Lample2016, Jaderberg2019, Berner2019, Vinyals2019}. 
Something these games have in common is existence of an API allowing researchers to interface with the game easily, and the ability to be simulated at speeds far quicker than real time and/or run it cheaply at scale. This is necessary for today's deep reinforcement learning (RL) algorithms, which require large amounts of experience to learn effectively --  for instance, an actor-critic algorithm used over 10,000 years of experience to master Dota 2.





Games without APIs, that can't be run easily at scale, have received less research attention. Without access to mass-scale simulations, one is forced to explore more efficient algorithms. In this paper we take on such a challenge; building an agent for Counter-Strike: Global Offensive (CSGO), with no pre-existing API, and only modest compute resources (8$\times$GPUs for training, 1$\times$GPU at test time, and a single game terminal).


Released in 2012, CSGO is one of the world's most popular games in player numbers and audience viewing figures.
The computational requirements of CSGO are an order of magnitude higher than the FPS games previously studied. For instance, while Doom can be run at 7000 frames-per-second on a single CPU \citep{Wydmuch2019}, CSGO runs at 200 frames-per-second on a modern GPU.

CSGO's constraints preclude mass-scale on-policy rollouts, and demand an algorithm efficient in both data and compute, which leads us to consider behavioural cloning. Whilst prior work has applied this to various games, demonstration data is typically limited to what authors provide themselves. Playing repetitive game modes at low resolution means these datasets remain small (one to five hours -- section \ref{sec_relatedwork}), producing agents of limited skill-level.

Our work takes advantage of CSGO's popularity to record data from other people's play -- by joining games as a spectator and scraping screenshots and inferring actions. This allows us to collect a dataset an order of magnitude larger than in previous FPS works, 5.5 million frames or 95 hours. We use a two-stage approach; initially training a deep neural network on this large noisy dataset, then fine-tuning it on smaller clean expert demonstrations. The resulting agent can play the game with a skill-level around that of the medium-difficulty built-in bot (the rules-based AI available as part of CSGO), or equivalently, a casual human FPS gamer.



Whilst RL research often aims to maximise reward, we emphasise that this is not the exclusive objective of this paper  -- perfect aim can be achieved through simple geometry and accessing backend information about enemy locations (hacks and built-in bots exploit this). Rather, we aim to produce an agent that plays in a humanlike fashion, that is fun and  challenging to play with and against.



This paper makes several \textbf{contributions:} 
1) Provides a blueprint for building data and compute efficient agents for modern games. 
2) Proposes a two-stage behavioural cloning approach.
3) First major work on a modern FPS game, and largest-scale behavioural cloning effort in this genre.
4) Introduces the CSGO environment, and human demonstration datasets, to the AI community.



\begin{figure}[t!]
\begin{center}
\begin{minipage}{.45\textwidth}
	\includegraphics[width=1.0\columnwidth]{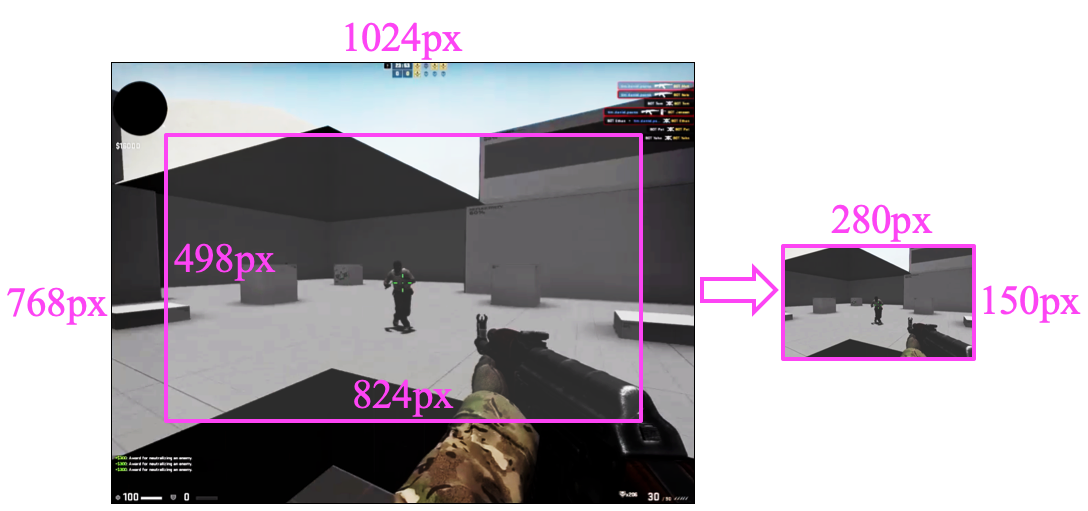}
	\caption{Screen processing involves cropping and downsampling. Aim training mode shown.}
	\label{fig_screenprocess}
\end{minipage}
\hspace{0.2in}
\begin{minipage}{.45\textwidth}
	\includegraphics[width=1.0\columnwidth]{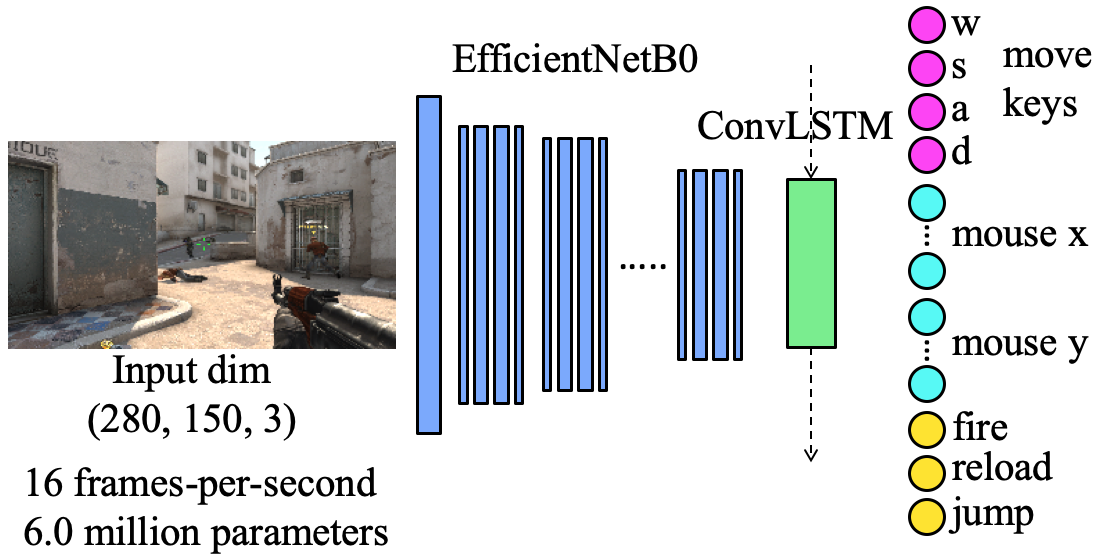}
	\caption{Overview of the agent's architecture. Deathmatch mode shown.}
	\label{fig_nn_architecture}
\end{minipage}
\end{center}
\end{figure}

\section{Background}
\label{sec_background}

This section describes the CSGO environment, and briefly outlines behavioural cloning.


\subsection{CSGO Environment}

CSGO is played from a first person perspective, with mechanics and controls that are standard across FPS games -- the keyboard is used to move the player left/right/forward/backwards, while mouse movement turns the player horizontally and vertically, serving both to look around and aim weapons. In CSGO's full `\textbf{competitive mode}', two teams of five players win either by eliminating the opposing team, or completing an assigned objective. Success requires mastery of behaviour at three time horizons; In the \textbf{short term} an agent must control its aim and movement, reacting to enemies. Over the \textbf{medium term} the agent must navigate through map regions, manage its ammunition and react to its health level. In the \textbf{long term} an agent should manage its economy, plan strategies, adapt to opponents' strengths and weaknesses and cooperate with teammates. 

As the first attempt to play CSGO from pixel input, we do not consider the full competitive mode. Instead we focus on two simpler modes, summarised in table \ref{tbl_modes} (full settings in appendix \ref{sec_app_running}). Screenshots for each mode can be found in figures  \ref{fig_screenprocess} (aim training) \& \ref{fig_nn_architecture} (deathmatch). All CSGO game modes are partially observable, stochastic environments.

`\textbf{Aim training mode}' provides a controlled environment for players to improve their aim, recoil control and reaction speed. The player stands fixed in the centre of a visually uncluttered map, while unarmed enemies run toward them. Players can not take damage, and ammunition is unlimited. This constitutes our simplest environment.


`\textbf{Deathmatch mode}' rewards players for eliminating any enemy on the opposing team (two teams, `terrorists' and `counter-terrorists'). After dying a player revives at a random location. Whilst it does not require the long-term strategies of competitive mode, other elements are intact. It is played on the same maps, with the full variety of weapons available. Ammunition must be managed, and the agent should distinguish between teammates and enemies. 

We consider three difficulty settings of deathmatch mode, all on `dust2' map, the agent on the terrorist team and `AK47' equipped. 1) \textbf{Easy} -- built-in bots on easy mode, bots use pistols, reloading not required, 12 vs 12. 2) \textbf{Medium} -- built-in bots on medium difficulty, any weapon, reloading required, 12 vs 12. 3) \textbf{Human} -- human players, any weapon, reloading required, 10 vs 10.

\begin{table}[t]
\begin{center}
\resizebox{0.6 \columnwidth}{!}{
\begin{tabular}{lccc}
    \toprule
    &  Short-term  & Medium-term  & Long-term  \\
    Game mode &  Reactive \& control & Navigation \& ammo & Strategy \& cooperation \\
   \midrule
    Aim training & \cmark & \xmark & \xmark \\
    Deathmatch   & \cmark & \cmark & \xmark \\
    Competitive    & \cmark & \cmark & \cmark \\
    \bottomrule
\end{tabular}
}
\end{center}
\caption{CSGO game modes and the behaviour horizon required for success in each.}
\label{tbl_modes}
\end{table}

\subsection{Behavioural Cloning}

In behavioural cloning (a form of `imitation learning') an agent learns to mimic the action, $\mathbf{a} \in \mathcal{A}$, a demonstrator would take given some observed state, $\mathbf{o} \in \mathcal{O}$. Typically the agent, parameterised by $\theta$, outputs a probability distribution over possible actions, ${\pi}_\theta(\hat{\mathbf{a}} | \mathbf{o})$.

Learning is based on a dataset of the demonstrator's behaviour. For $N$ such pairs, $\mathcal{D} = \{ \{\mathbf{o}_1, \mathbf{a}_1 \} \dots \{\mathbf{o}_N, \mathbf{a}_N \} \}$. In its vanilla form, behavioural cloning uses some loss function, $l: \mathcal{A} \times \mathcal{A} \to \mathbb{R}$ (e.g. cross-entropy or mean squared error), measuring the distance between predicted and demonstrated actions, and a model is trained to optimise,
\begin{align*}
    \theta = \text{argmin}_\theta \sum_i^N l\left( \mathbf{a}_i, {\pi}_\theta \left( \hat{\mathbf{a}}_i | \mathbf{o}_i \right) \right) .
\end{align*}



Behavioural cloning reduces learning a sequential decision making process to a supervised learning task. This can be a highly efficient method for learning \citep{bakker1993}, since an agent is told exactly how to behave, removing the challenge of exploration -- in reward-based learning an agent experiments to learn strategies by itself. 

One drawback is that the learnt policy can only perform as well as the demonstrator (and in practise may be worse since it is only an approximation of it). A second is that often only a small portion of the state space will have been visited in the demonstration dataset, but due to compounding errors, the agent may find itself far outside of this \citep{Ross2011, Laskey2017}  -- there is a distribution mismatch at test time, $p_\mathcal{D}(\mathbf{o}) \neq p_{\pi_\theta}(\mathbf{o})$.


\section{Agent Design}
\label{sec_design}
This section provides details of (and justification for) the major design decisions of the agent.

\subsection{Observation Space}


CSGO is typically run at a resolution of around 1920$\times$1080, which is larger than most GPUs can process at a reasonable frame rate, meaning downsampling was required. 
This gives rise to a trade-off between resolution, size of neural network, frames-per-second, GPU requirements and training dataset size. For instance, a lower resolution compromises the agent's skill in longer-range firefights but might allow a deeper neural network to run at more frames-per-second. Additionally, rather than using the whole screen, a central portion can be cropped, which provides higher resolution for the important region around the crosshair but at the cost of narrowing the field of view. 

In this work the game is run at 1024$\times$768 resolution, and the agent crops a central region of 824$\times$498, then downsamples it to 280$\times$150 (figure \ref{fig_screenprocess}). This allows state-of-the-art network architectures to run at 16 frames-per-second on an average gaming GPU.  



\textbf{Auxiliary Information.}
The cropped pixel region usefully excludes several visual artefacts which appear in spectator mode but not when actively playing. It also excludes the radar map, score feed, clock, health level and ammunition. We experimented providing some of these in vector form to the network, but found they were not critical to performance, and excluded them to simplify the design.




\subsection{Action Space}

The action space in CSGO is a mixture of discrete keys \& clicks and continuous mouse movements. To simplify learning, we restrict it's output space to actions essential for a reasonable level of play as per figure \ref{fig_nn_architecture}. It excludes other actions such as `walk' key -- appendix table \ref{tbl_action_space} details the full space. 

Success in firefights requires precise mouse movement (aiming) as well as coordination between actions (e.g. accuracy reduces when moving). This creates two main design challenges: 1) How to model the mouse movement space. 2) How to model actions that are not mutually exclusive (e.g. one might reload, jump and turn left simultaneously).




The agent paramerises \textbf{mouse movement} by changes in x \& y coordinates. Whilst it may seem natural to treat these as continuous targets, when combined with a mean squared error loss,  this led to undesirable behaviour in initial experiments (given a choice of two pathways, the agent would output a point midway between the two, which minimised mean squared error!) \citep{Ontanon2014}. Discretising the mouse space and framing it as a classification task was more successful. Our datasets include both discretised and continuous mouse data.

The discretisation itself required tuning and experimentation -- a finer grid allows more precise control but requires more data to train. 
We innovated an unevenly discretised grid, finer in the middle, and coarser at the edges -- it's more important for a player to be able to make fine adjusments when aiming, compared to when turning large angles, when precision matters less. This also reflects the histograms of mouse movements in human play which are roughly Gaussian distributed (appendix figure \ref{fig_stats_mouse}). The agent has 19 options for mouse x $\in \{-300, -200, ..., -10,-4,-2,0,2,4,10,.., 200, 300\}$, and 13 options for mouse y $\in \{-50, ..., -10,-4,-2,0,2,4,10,.., 50\}$. This reflects that vertical movements tend to be of lower magnitude than horizontal movement. 

To address the \textbf{mutually inclusive} (several actions can be applied simultaneously) nature of the action space, independent losses are used for each action -- binary cross entropy losses for keys and clicks, and multinomial cross entropy losses for each mouse axis. As such the agent outputs the \textit{marginal} distribution of each action rather than the \textit{joint} distribution, i.e. it assumes that each action is independent of all others. This simplifying assumption appears to work well enough in practise, though one could imagine specific situations it may be inappropriate -- for instance if choosing to step left and reload behind cover, or remain static and fire.
Future work could explore approaches like GAIL, providing an implicit joint distribution, or inputting selected actions in a recurrent manner.



\subsection{Neural Network Architecture}

The agent's architecture is summarised in figure \ref{fig_nn_architecture}. Many architectures designed for image classification make heavy use of pooling operations, which cause loss of spatial information. For our application, the location of objects within an image is important -- knowing that an enemy is present in the image is not enough, the agent must know its location to take action. As such, whilst an EfficientNetB0 \citep{Tan2019} forms the trunk of the network, only the first six residual stages are used -- for an input of 280$\times$150$\times$3, this outputs a feature map of dimension 18$\times$10$\times$112. The network is initialised with ImageNet weights.

The agent requires more than a single input frame to sense motion of itself and others. A stacked input approach, with several previous frames jointly fed into the network, was successful in aim training mode, but caused issues when navigating, with the agent often getting stuck in doors and corners. The final agent uses a convolutional LSTM layer \citep{Shi2015} after the EfficientNetB0, which largely remedies this, and also allows the possibility of longer-term memory. A linear layer connects the output layer. 


\subsection{Test Time}

The agent parameterises a probability distribution over each action independently. 
Combinations of actions may be applied at every time step.
At test time, each action can either be selected probabilisitically $\tilde{\mathbf{a}} \sim \pi_\theta(\hat{\mathbf{a}} | \mathbf{o})$, or according to the highest probability, $\tilde{\mathbf{a}} = \text{argmax}_{\hat{\mathbf{a}}} \pi_\theta(\hat{\mathbf{a}} | \mathbf{o})$. 

Selecting movement keys and mouse movement probabilisitically produced jerky, unnatural movement, so are selected via argmax. Certain actions -- reload, jump, fire -- seldom exceed the 0.5 threshold required to be chosen by argmax, so are selected probabilistically.



With actions being applied 16 times a second, mouse movement can appear stilted. In recorded gameplay demos, this is artificially increased to 32 by halving the mouse input magnitude and applying twice with a short delay. 





%
%
%
%
%
%
%

\section{Methods \& Data}
\label{sec_data}
This section introduces the methodology used for training the agent, describes collection of the demonstration datasets (summarised in table \ref{tbl_data}), and summarises some training details. Appendix \ref{sec_app_dataset} provides key stats and visualisations of the online dataset.


\subsection{Two-Stage Methodology}

One of the difficulties of using behavioural cloning in many applications is that sourcing a large dataset of demonstrations is generally time-consuming and/or costly. 

Much of the literature in video games manually records demonstrations by using a specially set up machine to log key presses and mouse movements, but this results in small datasets (repetitive game modes in low resolution are no fun!) and systems of limited performance. 




Prior work in Starcraft II showed that reasonable performance \textit{can} be achieved through behavioural cloning, provided one has access to a dataset of sufficient size \citep{Vinyals2019}. Whilst Vinyals et al. worked alongside game developer Blizzard, having access to a large dataset of logged states and actions, for many games such access is not possible. 


In lieu of such privileges, we developed a two stage method. In the stage one, we scrape a large dataset of human play from public online servers. We do not have access to the ground truth actions applied by the player, and instead build an inverse dynamics model to estimate these actions from metadata. This is used for pre-training a neural network. In stage two, we manually create small clean datasets that the network is fine-tuned on. The clean datasets have several advantages that drastically boost performance.
\begin{itemize}
\item Recording gameplay allows clean labelling of the actions. 
\item We restrict the player's action and observation space to match that of the agent‘s.
\item There are minor differences in the visuals rendered by the game when viewing players in spectator mode, compared to actively playing, e.g. red damage bar indicators are not displayed in the former.
\item The online dataset contains a large variety of play styles and equipment choices. The clean dataset allows the network to specialise to a single high-skill policy.
\end{itemize}
%
%
%

Combining two datasets in this way allows the agent to learn from the broad state-space coverage in the online dataset, without compromising on the quality of the final policy. The quantity of manual demonstrations required is an order of magnitude smaller than if exclusively trained on.

\begin{table}[b]
\begin{center}
\resizebox{0.7 \columnwidth}{!}{
\begin{tabular}{l c c c l l }
	\toprule
    Dataset abbreviation &  Frames & Hours & GB & Map, game mode & Source \\
   \midrule 
    Online    & 5,500,000 & 95 & 680 & Dust2, human deathmatch & Scraped online  \\
    Clean       & 190,000 & 3.3 & 24 & Dust2, various deathmatch & Clean expert demos \\
    Clean (Inferno)        & 10,000 & 0.18 & 1.2 & Inferno, medium deathmatch & Clean expert demos \\
    Clean (Mirage)        & 10,000 & 0.18 & 1.2 & Mirage, medium deathmatch & Clean expert demos \\
    Clean (Nuke)        & 10,000 & 0.18 & 1.2 & Nuke, medium deathmatch & Clean expert demos  \\
    Clean aim train        & 45,000 & 0.78 & 6 & Aim train mode & Clean expert demos  \\
    \bottomrule
\end{tabular}
}
\end{center}
\caption{Summary of all datasets used in training.}
\label{tbl_data}
\vspace{-0.2in}
\end{table}

\subsection{Large-Scale Online Demonstrations}

This dataset was scraped from official Valve servers by joining in spectator mode, and running a script both to capture screenshots (processed as in figure \ref{fig_screenprocess}) and metadata at 16 frames-per-second (see appendix \ref{sec_app_interfacing} for interfacing details). Note the naming of this dataset as `online' refers to the source being online Valve game servers, and \textit{not} to the offline/online learning paradigms in RL. 

The script tracked the current best performing player in the server in an attempt to collect higher-skill demonstrations.
Periods of player immobility were filtered out in post-processing.

\subsubsection{Action Inference with Inverse Dynamics Model}

Metadata does not contain the actions that were applied by the player. Rather, it contains information about the player state (e.g. weapon selected, available ammunition, health, score), position on map ($x,y,z$ coordinates), velocities, and orientation (roll and yaw). 


We developed a rules-based algorithm for the inverse dynamics model. Whilst some actions were straightforward to infer (e.g. firing is detected if ammunition decreased between two time steps). Others required testing and tuning. For instance, inferring keys moving a player forward/backwards/left/right, is an ill-posed problem -- there can be many other reasons for a change in velocity, such as weapon switching (heavy weapons makes players move slowly), bumping into objects, or taking damage. There were also inconsistent time lags between an action's application, its manifestation in the metadata, and observing the change on screen. \href{https://github.com/TeaPearce/Counter-Strike_Behavioural_Cloning}{\textcolor{blue}{See script}} \verb|dm_infer_actions.py| for details.

We tuned the inverse dynamics model until it was able to infer actions in most scenarios tested. Whilst this required significant effort in reverse-engineering, the value was in its scalability -- once written it could scrape gameplay continuously for days at a time, providing a quantity and variety of demonstrations that we couldn't produce manually.

\subsection{Clean Expert Demonstrations}



We created five clean datasets using a terminal set up to precisely log actions and take screenshots. We used a strong human player to provide the data (top 10\% of CSGO players, `DMG' rank). This player was only allowed to use actions the agent can output. The game was run at 1024$\times$768 resolution. The audio was muted and radar covered up to mimic the agent's observation space.
For some of the demonstrations in easy and medium mode, we slowed the game to half speed (dropping the capture rate accordingly) to further improve the quality of the demonstrations.



\subsection{Training Details}


The agent is initially trained on the online dataset (validating on medium deathmatch mode every two epochs). From this pre-trained checkpoint, fine-tuned versions were created by further training on one of the the clean expert datasets (validating on the relevant map and mode every four epochs). A batchsize of 4 and sequence length of 96 frames (6 seconds) were used (LSTM states are reset between each sequence). Data augmentation was applied to image brightness and contrast, but not to spatial transformations, since this would invalidate mouse labels. 
In addition to the losses discussed, the agent outputs and optimises a value function estimate ($v_{t} = r_t + \gamma v_{t+1}$, where, $r_t = 1.0 \text{ kills}_t - 0.5 \text{ deaths}_t - 0.02 \text{ shoot}_t$). This may have the effect of providing extra supervision as an auxiliary task. In this paper the value function estimate is not used for any further purpose.



Ten different models were trained on the online dataset under various hyperparameter settings. Training for each model used 4$\times$ Titan X GPUs -- time for one epoch on the online dataset varied from 1 to 8 hours, dependent on the data subset and architecture used (appendix table \ref{tbl_ablation_subset_lstm}). Models trained for between 10 and 30 epochs. Fine-tuning on the clean (`dust2') dataset took 15 minutes per epoch, typically requiring 12 to 32 epochs.

\section{Related Work}
\label{sec_relatedwork}

Appendix section \ref{sec_app_relatedwork} provides a comprehensive literature review which we summarise here.


FPS games have proved useful environments for RL research. Two 1990's games have been packaged in convenient APIs. Beattie et al. \citeyearpar{Beattie2016} released DeepMind Lab, built around Quake 3 (originally 1999), and Kempka et al. \citeyearpar{Kempka2016} introduced VizDoom (originally 1993). These environments are basic in comparison to CSGO (originally 2012), using low resolution textures, a smaller action space, and orders of magnitude less compute -- e.g. VizDoom allows simulation at 7000 frames-per-second on a single CPU core \citep{Wydmuch2019}, whilst CSGO runs at around 200 frames-per-second on a modern GPU.

These FPS environments have attracted much research, the majority applying standard reward-based learning such as actor-critic methods or Q-learning, e.g. \citep{Lample2016, Jaderberg2019}. Several efforts have trialled behavioural cloning, with demonstration datasets of around one hour \citep{Gorman2007, Harmer2018, Kanervisto2020}. Our work stands out both as the first to tackle a \textit{modern} FPS, and as the largest-scale effort in behavioural cloning.

Imitation learning has been explored in other genres most often at small scale (1-5 hours -- table \ref{tbl_prior_imitation}), with authors recording demonstration data themselves. There are several notable exceptions; Go (30 million frames) \citep{Silver2016}, Starcraft II (971,000 replays) \citep{Vinyals2019}, and Minecraft \citep{Guss2019} (500 hours). Using these larger datasets reasonable performance \textit{could} be achieved (e.g.  Vinyals et al.'s agent acheived a rank in the top 16\% of human players), and they have inspired much follow up work. We hope the dataset we contribute in the CSGO environment will be valued similarly.



 
The computational difficulty of generating on-policy data for CSGO makes the blossoming field of offline RL \citep{Levine2020} very relevant, where there has been recognition that leveraging existing datasets for tasks typically tackled through pure RL could greatly improve efficiency. Benchmarks and datasets in offline RL are often algorithmically generated \citep{Fu2020} -- this has found particular favour in Atari games \citep{Agarwal2020}. We hope our large-scale human demonstration dataset might find use in this field.


\section{Evaluation}
\label{sec_experiments}

This section evaluates the agent in three ways: 1) Measuring the score it achieves relative to both human players and the built-in rules-based bot. 2) Assessing the `humanlike-ness' of the agent's play style, done both qualitatively and also by quantitatively analysing its map coverage. 3) Measuring its ability to generalise to new maps.

Gameplay examples are shared at: \textcolor{blue}{\url{https://youtu.be/KTY7UhjIMm4}}. The first demos were selected by running the strongest agent on each mode and setting for five minutes, and selecting a one minute segment from each which showcases the agent's skill. 
The video further includes: illustrations of the the agent's common failures,
a longer unedited clip of the agent on the medium setting,
a longer unedited clip of the agent navigating in an empty map.
Code to run the agent is provided at: \textcolor{blue}{\url{https://github.com/TeaPearce}}.







\subsection{Hyperparameter Search}



We trained multiple versions of the agent on the online dataset varying several hyperparameters: 1) Adding an extra LSTM layer (256 units) after the convolutional LSTM. 2) Applying dropout to recurrent connections. 3) Training on different subsets of the data; as well as training over the full dataset, we also considered only sequences where the AK47 was equipped, and only sequences where the AK47 or M4A1 was equipped. 4) We optionally undersampled sequences where a player did not score a kill (`non-scoring').

Appendix \ref{sec_app_results} provides numerical results in the medium difficulty deathmatch mode for various hyperparameters (note models typically take 2-5 days to train \& test, so not all permutations could be considered). In general, the extra LSTM layer was harmful, while undersampling with a probability of 0.2 or 0.4 was helpful, as was adding dropout to recurrent connections (dropconnect was always used in convolutional layers of the network). Notably, training on the full dataset was \textit{worse} than when training only on those sequences using AK47 (28\% of online dataset) or AK47 \& M4A1 (both are similar weapons, together making 40\% of online dataset). We believe this is due to different equipment requiring different play styles, for example players with sniper rifles (20\% of online dataset) are typically less mobile, and demand different aiming mechanics -- including this kind of data seems to degrade the ability of the agent with the AK47.







\subsection{Agents \& Baselines}




Subsequent analysis and results use the strongest version of the agent unless otherwise stated. This strongest agent was trained only on AK47 data, undersampling non-scoring sequences with a probability of 0.4, with recurrent dropout, and no extra LSTM layer. This is termed `\textbf{online agent}'.
This online agent was then fine-tuned on the clean datasets, producing `\textbf{fine-tuned dm agent}' and `\textbf{fine-tuned aim agent}'.





We include several baselines.  
1) \textbf{Built-in Bot (easy)} -- the bots played against in the deathmatch easy setting.
2) \textbf{Built-in Bot (medium)} -- the bots played against in the deathmatch medium setting.
3) \textbf{Human (Non-gamer)} -- someone with little experience playing games.
4) \textbf{Human (Casual gamer)} -- a regular player of video games, with a small amount ($<$100 hours) of CSGO experience.
5) \textbf{Human (Strong)} -- a player ranked in the top 10\% of regular CSGO players (`DMG' rank). All humans play at full 1920$\times$1080 resolution, and are assessed over 5 minutes (aim train mode) and 10 minutes (deathmatch modes) of play. Longer periods resulted in fatigue and decreased performance.


\begin{table*}[t!]
\vspace{-0.1in}
\begin{center}
\resizebox{0.9 \textwidth}{!}{
\begin{tabular}{l c cc cc cc c}
    \toprule
    & & & \multicolumn{6}{c}{--------------------------------------- Deathmatch ---------------------------------------} \\
    &  \multicolumn{1}{c}{Aim Train} & & \multicolumn{2}{c}{  ----- Easy  ----- } & \multicolumn{2}{c}{----- Medium -----} & \multicolumn{2}{c}{----- Human -----}  \\
   & KPM &  & KPM & K/D & KPM & K/D & KPM & K/D \\ 
     \midrule
     \textbf{Dataset used}  & & & & & & \\
	
	Online dm & 4.31 $\pm$ 0.20 &  &   3.47 $\pm$ 0.12 & 2.70 $\pm$ 0.26   &  2.23 $\pm$ 0.26 & 1.04 $\pm$ 0.06 &   0.68 $\pm$ 0.13 & 0.22 $\pm$ 0.03 \\
	Online dm + Clean aim train & 26.86 $\pm$ 0.39 &  &  -- & --  &   -- & -- &   -- & -- \\
	Online dm + Clean dm & -- &  &  5.06 $\pm$ 0.31 & 3.87 $\pm$ 0.24   &   3.72$\pm$0.25 & 2.09$\pm$0.19  &   1.43 $\pm$ 0.17 & 0.59 $\pm$ 0.09 \\

%

    
    \hdashline 
    \textbf{Baselines} & & & & & & 
    
    \\
    Built-in Bot (easy) & -- &  &  2.11 & 1.00 &  -- & -- &  -- & -- \\
    Built-in Bot (medium) & -- &  &  -- & --  &   2.41 & 1.00 &   -- & -- \\
    Human (Non-gamer) & 14.32 &  &  4.25 & 1.80  & 2.38 & 0.90 &   0.75 & 0.27 \\
    Human (Casual gamer) & 26.21 &  &  4.20 & 4.20  &   3.51 & 2.48 &  1.64 & 0.64 \\
    Human (Strong CSGO player) & 33.21 &  &  14.00 & 11.67  &   7.80 & 4.33 &   4.27 & 2.34 \\
    \bottomrule
\end{tabular}
}
\caption{Main results. Metrics are kills-per-minute (KPM) and kills/death ratio (K/D). Higher is better. Mean $\pm$ one std. error.}
\vspace{-0.2in}
\label{tbl_performance}
\end{center}
\end{table*}

\subsection{Main Results}

Table \ref{tbl_performance} displays results of the strongest agent. We report two metrics; kills-per-minute (KPM) and kill/death ratio (K/D). A strong player should have a high KPM and high K/D. For each game mode and setting we report the mean and one standard error over eight episodes of 10 minutes. The skill of bots and humans can vary from server to server, so we restart the game at least three times within these eight episodes.


\textbf{Aim train mode:}\footnote{K/D is not applicable to this mode since the agent never dies.}
The fine-tuned aim agent's performance is in line with the casual gamer's. The agent demonstrated good accuracy and recoil control, prioritising enemy targets sensibly, and anticipating their motion. Aim train mode required less clean training data than deathmatch for good performance (45 minutes vs 190 minutes) -- this is because it is a visually simpler environment and requires behaviour over short time horizons only (table \ref{tbl_modes}).
The online agent was able to somewhat generalise to the new environment, although it was unfamiliar with the specific movement patterns of enemies.





\textbf{Deathmatch mode:} The fine-tuned dm agent outperforms the built-in bot, both on easy and medium settings, roughly matching the performance of the casual gamer. 
The agent navigates around the majority of the map well, identifying and reacting to enemies reliably, and distinguishing them from teammates. It also chooses sensible moments to reload.
Moving from the online agent to the fine-tuned agent improves KPM by around 45\%, showing the importance of converging on a single high-skill policy. However, a performance gap remains between the agent and strong human.

\subsection{Humanlike Assessment}


As mentioned before, this paper is not exclusively interested in the score-based performance of the agent. Another key goal of agent design in video games it to build humanlike agents.
One of the advantages of using behavioural cloning is that the agent naturally adopts humanlike traits (even if sub-optimal!).
Measurement of this humanlike quality is less straightforward, and we qualitatively discuss traits observed during testing. The agent's map coverage is then quantitatively analysed.



\textbf{Humanlike traits:} 
The agent's mouse movement mimics that of a human, pausing mid-turn as if the mouse had reached the edge of the mouse pad. 
The agent's navigation is quite humanlike, often running along ledges or jumping over obstacles. In certain areas it will jump to spot an occluded area. The online agent sometimes exhibits playful quirks, such as firing at chickens or `bunny-hopping'.
The fine-tuned dm agent occasionally employs higher-skill behaviours, such as moving behind cover when reloading, or strafing during fire-fights -- the version of the agent trained on AK47 \& M4A1 sequences does this more often than the strongest AK47-only agent.

\textbf{Non-humanlike traits:} 
The agent makes several mistakes humans do not. It's memory is poor -- if an enemy disappears behind cover it quickly forgets about it. It also does not pick up on `second-order clues' about where enemies may be (e.g. teammates firing in some direction). It is poor at aiming vertically. In one region of the map that is rarely visited by human players (bottom left in figure \ref{fig_map_coverage_small}) its navigation is poor. Occasionally (once in 10 minutes) the agent gets into a position it can't recover from.



There are several more understandable limitations. The agent only receives the image as input, so has no audio clues that humans react to (shots being fired, or enemy footsteps). It also rarely reacts to red damage bar indicators, since these are not displayed in the online dataset. 





\subsubsection{Quantitative Map Coverage Analysis}

To quantitatively assess the similarity of the agent to human play, we track the $x$ \& $y$ coordinates of the agent playing on the medium deathmatch mode for 100 minutes. 
We discretise the map on a 60$\times$60 grid, and count the amount of time spent in each box. This distribution is compared to human play in the online and clean datasets, as well as the built-in bot.

\begin{figure}[h!]
\vspace{-0.1in}
\begin{center}
\includegraphics[width=0.2\columnwidth]{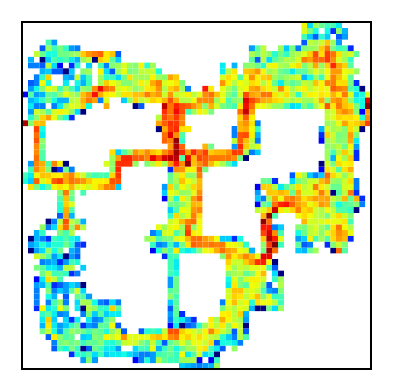}
\put(-65,-4){\small Online dataset}
\includegraphics[width=0.2\columnwidth]{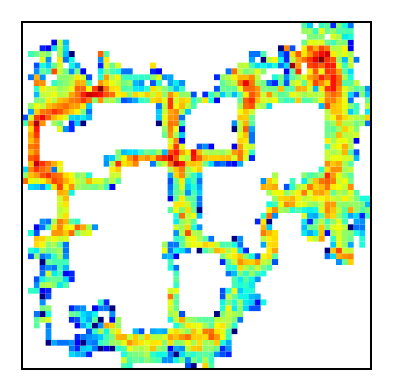}
\put(-70,-4){\small Online dm agent}
\includegraphics[width=0.2\columnwidth]{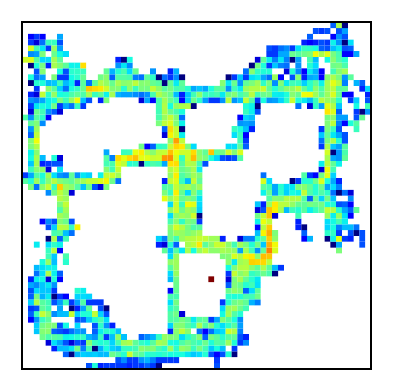}
\put(-75,-4){\small Medium built-in bot}

\includegraphics[width=0.2\columnwidth]{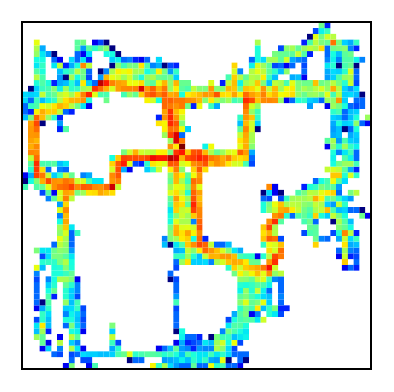}
\put(-70,-4){\small Clean dm dataset}
\includegraphics[width=0.2\columnwidth]{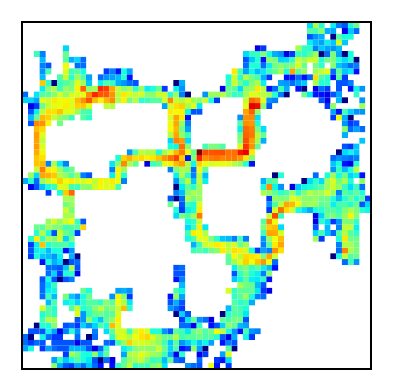}
\put(-75,-4){\small Fine-tuned dm agent}
\includegraphics[width=0.2\columnwidth]{images/03_map_coverage/map_coverage_bot.pdf}
\put(-75,-4){\small Medium built-in bot}
\caption{Map coverage heatmap for agent, built-in bot, and human datasets. Note the agent's coverage mimics the data it's trained on. Quantitatively this is captured by the `Earth mover's distance' between each distribution as in table \ref{tbl_earth_movers}.}
\label{fig_map_coverage_small}
\vspace{-0.2in}
\end{center}
\end{figure}

We consider two versions of the agent; one trained over the \textit{full} dataset and one subsequently fine-tuned on the clean dm dataset. Figure \ref{fig_map_coverage_small} shows map coverage heatmaps for each policy. One first observes that the agent's histograms mimic the routes taken by human players more closely than the built-in bots. Secondly, the online dm agent's coverage is most similar to that of the online dataset, while the coverage of the agent fine-tuned on the clean dataset is most similar to the clean dm dataset. These observations are quantified in table \ref{tbl_earth_movers}, where similarity between pairs of distributions is computed via the Earth mover's distance.

\begin{table}[h!]
\begin{center}
\resizebox{0.6 \columnwidth}{!}{
\begin{tabular}{l ccccc}
\toprule
     & Online dm & Clean dm & Online agent (full) & Fine-tuned agent & Built-in Bot \\
  \midrule 
    Vs. Online dm & 0.000 & 0.977 & \textbf{0.447} & 0.941 & 0.624 \\
    Vs. Clean dm & 0.970 & 0.000 & 0.533 & \textbf{0.349} & 0.643 \\
    \bottomrule
\end{tabular}
}
\end{center}
\caption{Map coverage analysis showing similarity between trajectories (100 minutes) from different pairs of datasets/policies. Earth mover's distance, lower is more similar.}
\label{tbl_earth_movers}
\vspace{-0.2in}
\end{table}

\subsection{Zero and Few-Shot Generalisation}

This section tests the ability of the agent to generalise to new environments under the deathmatch game mode. These have different layouts, appearances (e.g. `dust2' is set in a sandy Moroccan environment, `nuke' uses a disused power plant), and player models. Equipment and team is unchanged.  Appendix figure \ref{fig_maps_generalise} provides visualisations of each.

We first take the fine-tuned dm agent as reported in table \ref{tbl_performance}, trained only on the `dust2' map, and roll it out directly in the new environment (zero-shot). We then fine-tune the model on a small amount (10 minutes) of training data collected in the new environment (as per table \ref{tbl_data}), and retest it (few-shot). Results in table \ref{tbl_newmaps} are over five episodes of 10 minutes.

The agent shows some zero-shot generalisation ability in completely new environments, scoring 0.2-0.5 KPM, which improves to 0.7-1.0 with 10 minutes of training data. A useful comparison point is given by the ablation where an agent is trained from scratch on `dust2' (appendix figure \ref{fig_ablation_expertdata_pretrain}), scoring around 0.5 and 1 KPM after using 1 and 2 hours of clean expert data. Hence, the agent can significantly save on data collection in new environments. 


\begin{table}[h!]
\begin{center}
\vspace{-0.05in}
\resizebox{0.5 \columnwidth}{!}{
\begin{tabular}{l cc cc}
\toprule
    & \multicolumn{2}{c}{-- Zero-shot --} & \multicolumn{2}{c}{-- Few-shot --}  \\
    Map & KPM & K/D & KPM & K/D  \\
  \midrule 
    Inferno & 0.44 $\pm$ 0.09 & 0.21 $\pm$ 0.04 & 0.84 $\pm$ 0.13 & 0.45 $\pm$ 0.09 \\
    Mirage & 0.46 $\pm$ 0.05 & 0.25 $\pm$ 0.03 & 0.71 $\pm$ 0.20 & 0.31 $\pm$ 0.08  \\
    Nuke & 0.19 $\pm$ 0.05 & 0.13 $\pm$ 0.04 & 1.04 $\pm$ 0.12 & 0.52 $\pm$ 0.05  \\
    \bottomrule
\end{tabular}
}
\end{center}
\caption{Generalisation to new maps. Mean $\pm$ one std. error.}
\label{tbl_newmaps}
\vspace{-0.1in}
\end{table}

\subsection{Ablations on Data Size and Pre-training}

Ablations were run to study the benefit of pre-training on the online dataset, compared with training models from scratch on the clean deathmatch dataset. These experiments were applied to the agent trained over the \textit{full} online dataset.
Using three choices of weight initialisation, \{random, ImageNet, online dataset pre-training\}, we fine-tuned on varying amounts of clean demonstration data; \{1, 2, 3\} hours. Results are shown in appendix figure \ref{fig_ablation_expertdata_pretrain}.
Whilst using the weights from ImageNet improved over random initialisations, there is a large benefit to using weights pre-trained on the online dataset -- extrapolating the curves in the figure suggest around 20 hours of clean expert data would be needed to match the performance of the pre-trained model fine-tuned on just 2 hours of clean expert data.

\section{Discussion \& Conclusion}
\label{sec_discussion_conclusion}

This paper presented an AI agent that plays the video game CSGO from pixels, matching the skill-level of a casual human gamer. It is among the first efforts to tackle a modern video game, and the largest-scale work in behavioural cloning in the FPS genre to date.



Whilst the AI community has historically focused on real-time-strategy games such as Starcraft, we see CSGO as an equally worthy test-bed, providing its own unique mix of control, navigation, teamwork, and strategy. Its large and long-lasting player base, as well as similarity to other FPS titles, means AI progress in CSGO is likely to attract broad interest, and also suggests tangible value in developing strong, humanlike agents.

Although an inconvenience to researchers, CSGO's inability to be simulated at scale arguably creates a challenge more representative of those in the real-world, where RL algorithms can't always be run from a blank state. As such, CSGO lends itself to offline RL research. This paper has defined several game modes of varying difficulty, and had a first attempt at solving them with behavioural cloning. We share our code and datasets to encourage other researchers to partake in this environment's challenges.



There are many directions in which our research might be extended, such as applying more advanced methods from imitation learning or offline RL, or integrating with reward-based learning.
More ambitiously, there's the challenge of taking on CSGO's full competitive mode -- we see our paper as a step toward that AI milestone.






\section*{Ethical Considerations \& Broader Impact}
This work takes place in a simulated environment containing military themes, which may raise concerns around the usage of AI and weapons. This work purely aims to provide research in the domain of video games, of which the first-person-shooter is a widely popular genre, and not with any weapon development agenda in mind. Related research using the Doom and Quake engines has proven to be useful to the research community without (to the best of or knowledge) finding use in development of weapons. Nevertheless, we take this opportunity to remind the community to be cognisant of this risk as the gap between video games and reality narrows in future.


Regarding data privacy and legal issues, we have taken steps to anonymise the datasets, excluding player handles from the metadata.
We have communicated directly with CSGO's developer Valve about the sharing of the datasets and code from this paper. They have approved the use of these in a research context.


\section*{Acknowledgments}
Thanks to Bang Xiang Yong, Grace E. Lee and Charlotte Morrill for providing the baselines. Also to Sam Devlin for discussion, and anonymous reviewers for feedback that has improved the paper. 

\bibliography{library.bib}

\newpage
\appendix

\onecolumn

\begin{center}
\textbf{\Large{Supplementary Material: Counter-Strike Deathmatch with Large-Scale Behavioural Cloning}}
\end{center}

The supplementary material contains the following sections:
\begin{itemize}
    \item \textbf{Section \ref{sec_app_dataset}, Dataset Details:} High-level statistics and visualisations of the large dataset. Example image sequences.
    \item  \textbf{Section \ref{sec_app_results}, Further Results:} Ablations and further results.
    \item \textbf{Section \ref{sec_app_relatedwork}, Comprehensive Related Work:} Expanded related work section.
    \item  \textbf{Section \ref{sec_app_interfacing}, Interfacing with the Game:} Details of packages used to interact with the game.
    \item  \textbf{Section \ref{sec_app_running}, CSGO Game Settings:} Console commands used for each environment tested.
    \item  \textbf{Section \ref{sec_app_misc}, Miscellaneous:} Description of CSGO's full action space.
\end{itemize}

\newpage

\section{Dataset Details}
\label{sec_app_dataset}

\subsection{Key Statistics}

Key statistics from the online dataset.
\begin{itemize}
\item Total frames 5,500,000
\item Total kill events 26,516
\item Total death events 16,324
\item Number of AK47 frames 1,399,942
\item Number of AK47 kill events 8,344
\end{itemize}

\begin{figure}[h!]
\begin{center}
\includegraphics[width=0.5\columnwidth]{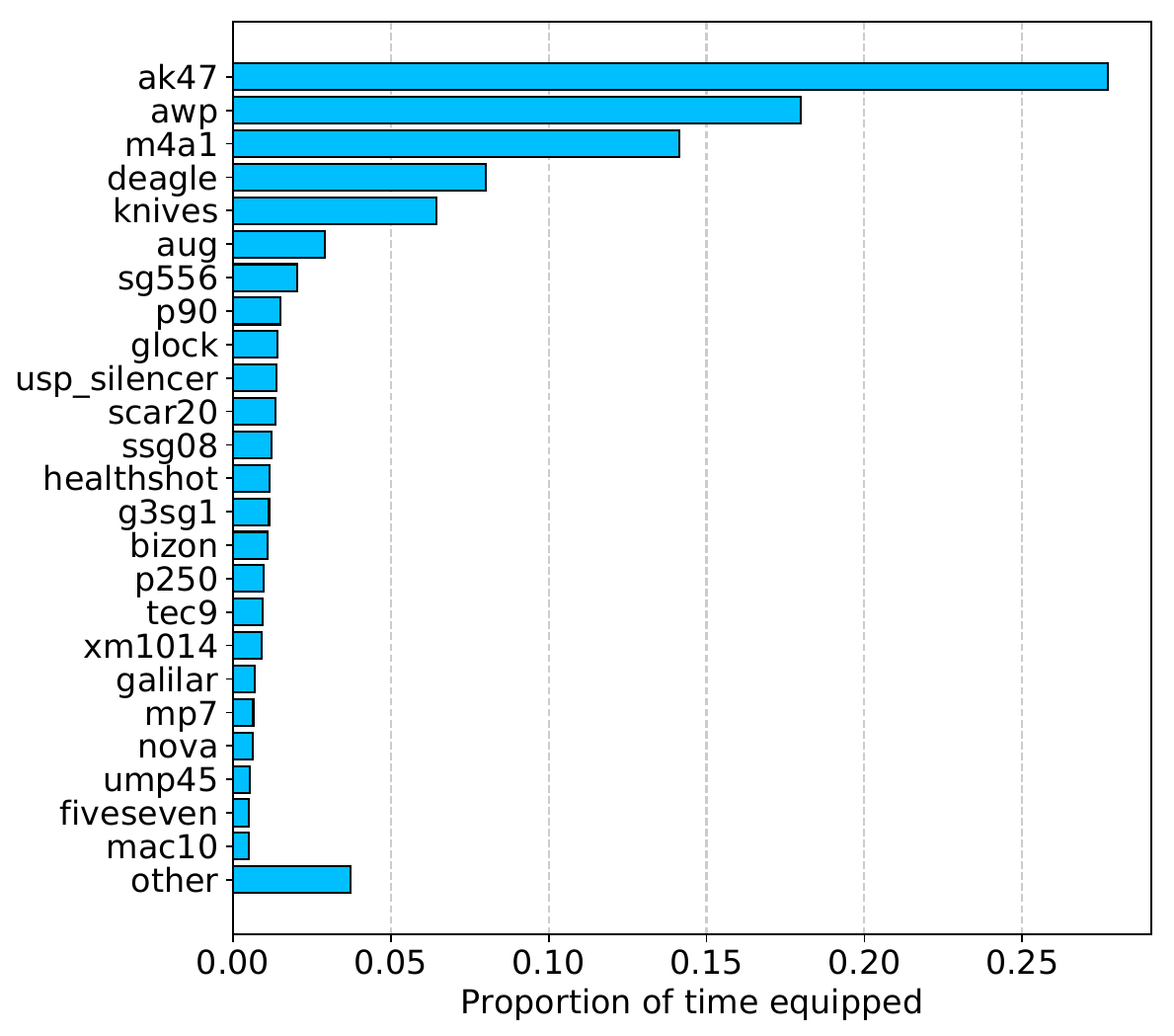}
\vskip -0.1in
\caption{Proportion of time each item is equipped in the online dataset.}
\label{fig_stats_equip}
\vskip -0.1in
\end{center}
\end{figure}

\begin{figure}[h!]
\begin{center}
\includegraphics[width=0.45\columnwidth]{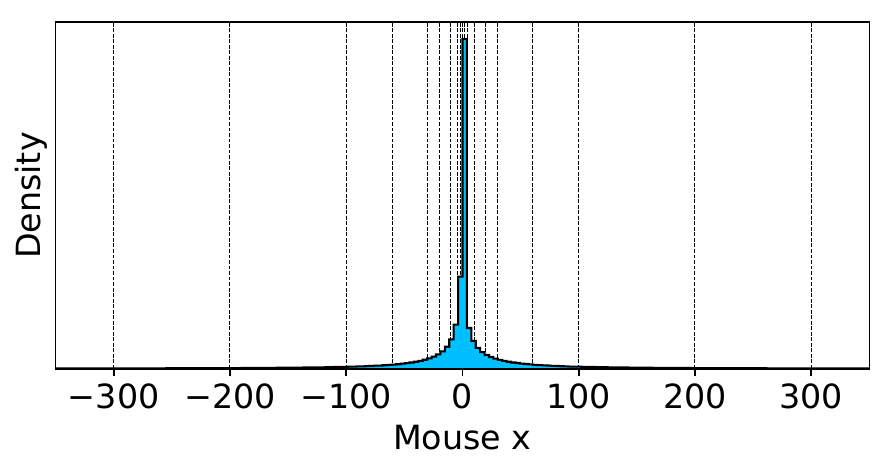}
\includegraphics[width=0.45\columnwidth]{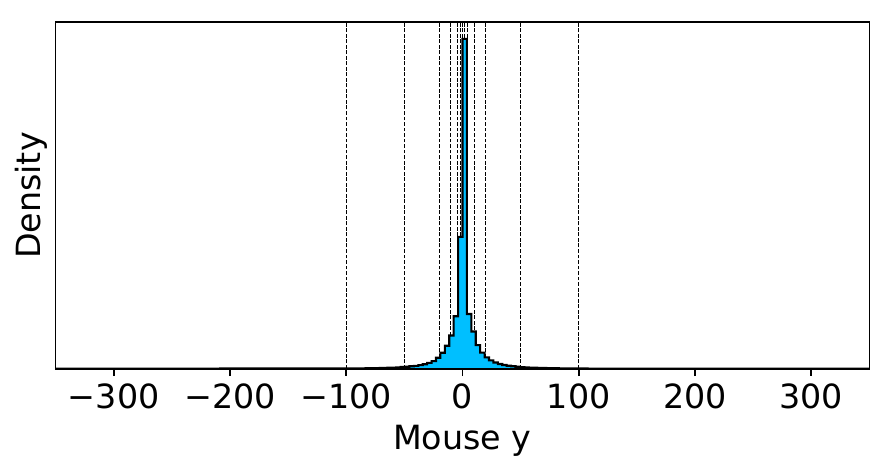}
\vskip -0.1in
\caption{Mouse movement histograms in the online dataset. Note mouse x has a larger variance than mouse y. Dashed lines overlaid show discrete options output by the agent.}
\label{fig_stats_mouse}
\vskip -0.1in
\end{center}
\end{figure}

\newpage
\subsection{Visualisation}

\vspace{1in}
\begin{figure}[h!]
\begin{center}
\includegraphics[width=0.245\columnwidth]{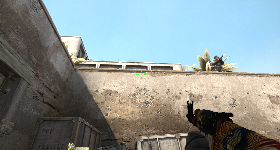}
\includegraphics[width=0.245\columnwidth]{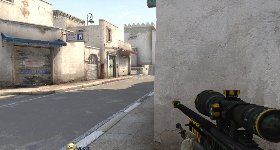}
\includegraphics[width=0.245\columnwidth]{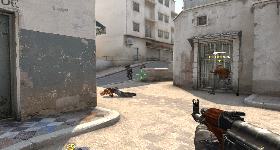}
\includegraphics[width=0.245\columnwidth]{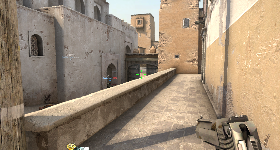}

\includegraphics[width=0.245\columnwidth]{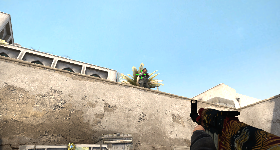}
\includegraphics[width=0.245\columnwidth]{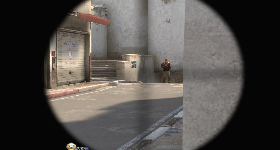}
\includegraphics[width=0.245\columnwidth]{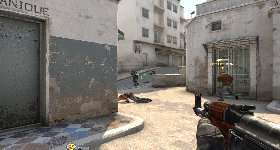}
\includegraphics[width=0.245\columnwidth]{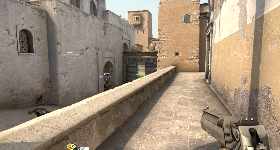}

\includegraphics[width=0.245\columnwidth]{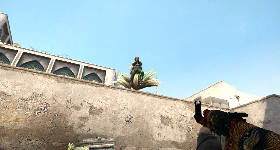}
\includegraphics[width=0.245\columnwidth]{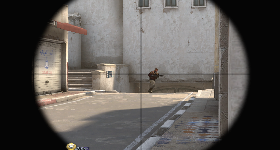}
\includegraphics[width=0.245\columnwidth]{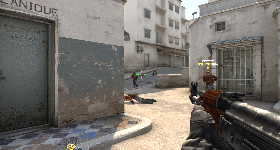}
\includegraphics[width=0.245\columnwidth]{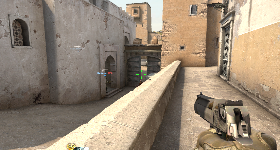}

\includegraphics[width=0.245\columnwidth]{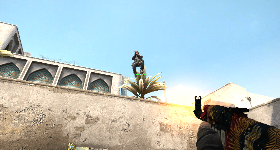}
\includegraphics[width=0.245\columnwidth]{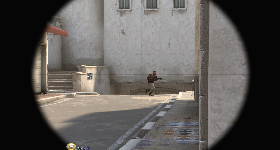}
\includegraphics[width=0.245\columnwidth]{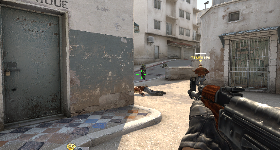}
\includegraphics[width=0.245\columnwidth]{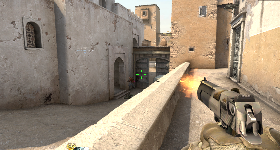}

\includegraphics[width=0.245\columnwidth]{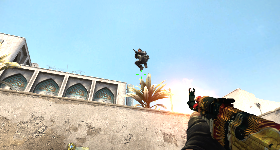}
\includegraphics[width=0.245\columnwidth]{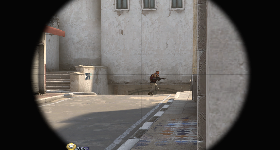}
\includegraphics[width=0.245\columnwidth]{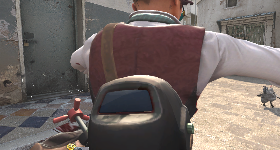}
\includegraphics[width=0.245\columnwidth]{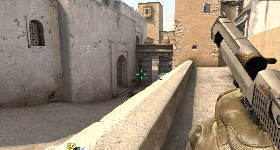}

\includegraphics[width=0.245\columnwidth]{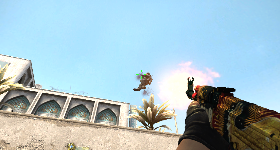}
\includegraphics[width=0.245\columnwidth]{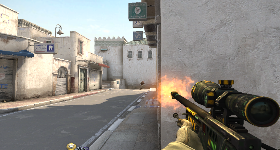}
\includegraphics[width=0.245\columnwidth]{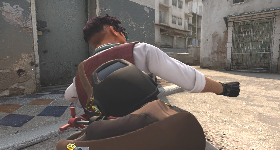}
\includegraphics[width=0.245\columnwidth]{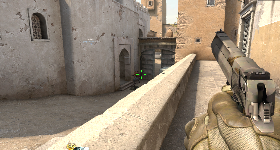}

\caption{Example image sequences from the online dataset.}
\label{fig_training_egs}
\end{center}
\vskip -0.1in
\end{figure}

\begin{figure}[h!]
\begin{center}
\includegraphics[width=0.6\columnwidth]{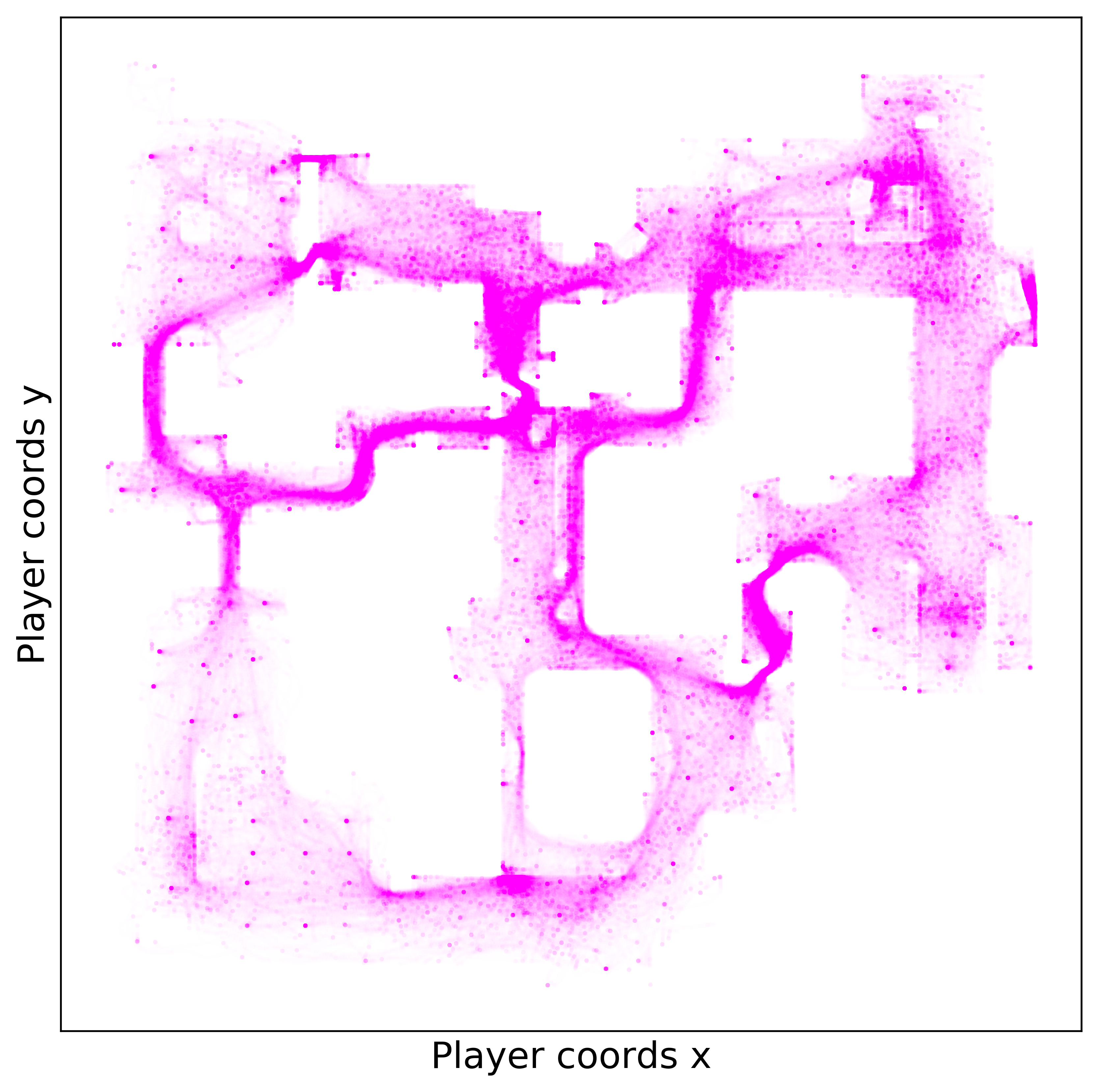}
\vskip -0.1in
\caption{Player trajectories over 30 hours of the online dataset.}
\label{fig_stats_trajectories}
\vskip -0.1in
\end{center}
\end{figure}



\begin{figure}[h!]
\begin{center}
\includegraphics[width=1.\columnwidth]{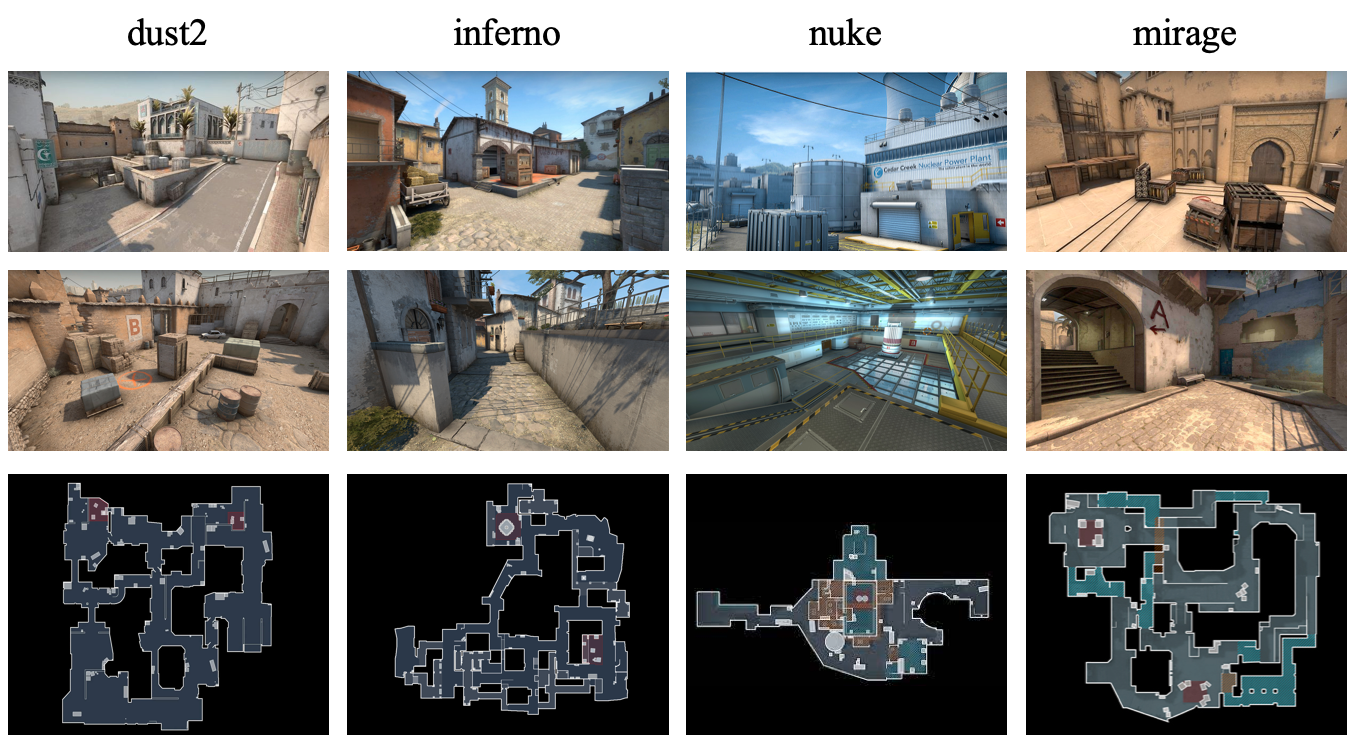}
\vskip -0.1in
\caption{Visual comparison of various CSGO maps used in generalisation testing -- scenic shots and radar overview.}
\label{fig_maps_generalise}
\vskip -0.1in
\end{center}
\end{figure}

%

\FloatBarrier

\newpage
\section{Further Results}
\label{sec_app_results}

\begin{figure}[h!]
\begin{center}
\includegraphics[width=0.4\columnwidth]{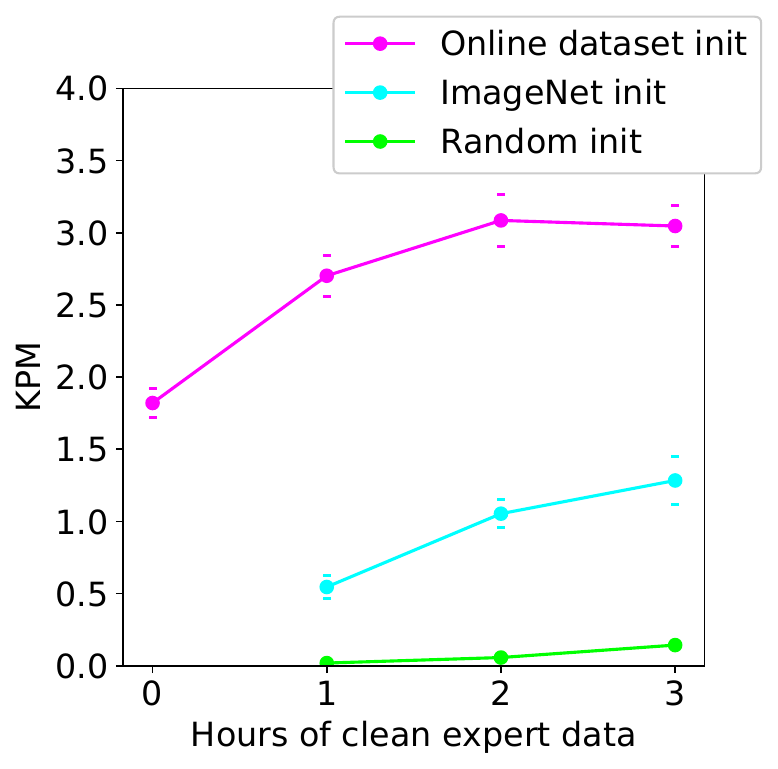}
\vskip -0.1in
\caption{Ablation of initialisation point and amount of clean expert data.  Medium difficulty deathmatch over 10 episodes of 10 minutes. Mean $\pm$ one standard error. Note that the agent trained over the full dataset was used for the `online dataset init' -- the performance gap would be even larger with other strong versions.}
\label{fig_ablation_expertdata_pretrain}
\vskip -0.1in
\end{center}
\end{figure}

\vfill
\begin{table}[h!]
\begin{center}
\resizebox{0.95 \columnwidth}{!}{
\begin{tabular}{l cc cc}
\toprule
    & \multicolumn{2}{c}{EffNetB0 + ConvLSTM} & \multicolumn{2}{c}{EffNetB0 + ConvLSTM + LSTM}  \\
    Data subset \& details & KPM & K/D & KPM & K/D  \\
  \midrule 
  \multicolumn{4}{l}{\textbf{Pretrained on online dataset}} \\
    AK47 & 1.68 $\pm$ 0.10 & 0.7 $\pm$ 0.07 & -- & --\\
    AK47 undersampled, p=0.4 & 2.03 $\pm$ 0.11 & 0.83 $\pm$ 0.05 & 1.88 $\pm$ 0.13 & 0.75 $\pm$ 0.06 \\
    AK47 undersampled, p=0.4, w/ recurrent dropout & 2.23 $\pm$ 0.26 & 1.04 $\pm$ 0.06 & 1.99 $\pm$ 0.15 & 0.92 $\pm$ 0.05 \\ 
    AK47 + M4A1 & 2.16 $\pm$ 0.18 &  0.84 $\pm$ 0.10 & 1.82 $\pm$ 0.24 & 0.75 $\pm$ 0.11\\
    AK47 + M4A1 undersampled, p=0.2 & 2.01 $\pm$ 0.10 & 0.85 $\pm$ 0.06 & -- & -- \\ 
    AK47 + M4A1 undersampled, p=0.2, w/ recurrent dropout & 2.24 $\pm$ 0.14 & 0.98 $\pm$ 0.11 & -- & -- \\
    Full dataset & 1.61 $\pm$ 0.14 & 0.67 $\pm$ 0.05 & -- & -- \\
   \hdashline 
   
  \multicolumn{4}{l}{\textbf{Pretrained on online dataset then finetuned on clean expert dataset}} \\
    AK47 & 3.14 $\pm$ 0.19 & 1.37 $\pm$ 0.07 & -- & --\\
    AK47 undersampled, p=0.4 & 3.01 $\pm$ 0.30 & 1.84 $\pm$ 0.21 & 2.89 $\pm$ 0.12 & 1.36 $\pm$ 0.07 \\ 
    AK47 undersampled, p=0.4, w/ recurrent dropout & 3.72 $\pm$ 0.25 & 2.09 $\pm$ 0.19 & 3.57 $\pm$ 0.14 & 1.74 $\pm$ 0.09 \\ 
    AK47 + M4A1 & 3.17 $\pm$ 0.16 & 1.61 $\pm$ 0.12 & 3.19 $\pm$ 0.13 & 1.41 $\pm$ 0.06 \\
    AK47 + M4A1 undersampled, p=0.2 & 3.23 $\pm$ 0.22 & 1.64 $\pm$ 0.14 & -- & --  \\ 
    AK47 + M4A1 undersampled, p=0.2, w/ recurrent dropout & 3.32 $\pm$ 0.16 & 1.69 $\pm$ 0.15 & -- & -- \\
    Full dataset & 2.95 $\pm$ 0.10 & 1.55 $\pm$ 0.08 & -- & -- \\
    \bottomrule
\end{tabular}
}
\end{center}
\caption{Various version of the agent trained with various hyperparameters. Medium difficulty deathmatch over eight episodes of 10 minutes. Mean $\pm$ one standard error. Data subset refers to which part of the full online dataset was used for training. The bottom half of the table reports results after the corresponding model from the top half was finetuned on the clean dataset. 
Undersampled refers to undersampling sequences that do not contain a kill event with probability as given. 
We consider two architectures; EffNetB0 + ConvLSTM is as given in figure \ref{fig_nn_architecture}, and EffNetB0 + ConvLSTM + LSTM which is similar but adds an extra dense LSTM layer. All models used dropconnect in the convolutional trunk of the model, recurrent dropout refers to dropout turned on (p=0.5) for recurrent connections.}
\label{tbl_ablation_subset_lstm}
\end{table}

\newpage
\section{Comprehensive Related Work}
\label{sec_app_relatedwork}
\textbf{FPS games:} FPS games have been proposed as useful environments for RL research,
 some being packaged in convenient APIs. Beattie et al. \citeyearpar{Beattie2016} released DeepMind Lab, built on the Quake 3 engine (originally 1999), and Kempka et al. \citeyearpar{Kempka2016} introduced VizDoom, packaging several simple game modes on top of the Doom engine (originally 1993). These environments are basic in comparison to CSGO (originally 2012) -- these 1990's FPS games were designed to be played at low resolutions, and VizDoom provides a low dimensional mouse action space. As a concrete comparison, VizDoom allows simulation at 7000 frames-per-second on a single CPU \citep{Wydmuch2019}, whilst CSGO runs at around 200 frames-per-second on a modern GPU.

The VizDoom and DeepMind Lab environments have inspired much follow up research. Notably, in the latter environment, Jaderberg et al. \citeyearpar{Jaderberg2019} considered a capture the flag mode, and trained agents able to outperform teams of humans familiar with FPS games --  they used an actor-critic algorithm also trained on auxiliary tasks. The agent received an input of resolution 84x84 pixels, with a mouse output space of 5x3 (five options for mouse x and three for mouse y), run at 15 frames-per-second, and learnt over 2 billion frames. Tournament-style contests have been hosted on a deathmatch mode of VizDoom \citep{Wydmuch2019}, with the strongest agents using either actor-critic methods or Q-learning, and often using separate modules for movement and firing. Their performance was below human level. One simple way to improve performance of FPS agents is to add auxiliary tasks predicting game feature information (e.g. presence and location of enemies) in parallel to learning a policy \citep{Lample2016}. Improving decision making at longer time horizons has been investigated through hierarchical approaches and cleverly compressing the action space \citep{Song2019, Huang2019}.

There are two main differences between this prior FPS work and our own: 1) We consider a modern FPS game bringing several new challenges (no API, better graphics, larger action space). 2) We focus on a behavioural cloning approach.







\textbf{Imitation learning for video games:} Various authors have experimented building agents for games using imitation learning. We summarise some of this work in table \ref{tbl_prior_imitation}. Typically the datasets are created by the authors themselves, which results in rather small datasets (1-5 hours), and limited agent performance. A common approach is to use a policy trained on behavioural cloning as a warm start for other RL algorithms.

To our knowledge the largest behavioural cloning efforts in games to date are;
Go (30 million frames) \citep{Silver2016}, Starcraft II (971,000 replays) \citep{Vinyals2019}, and Minecraft \citep{Guss2019} (500 hours). 
Of these, only Minecraft is `from pixels' -- as Berner et al. \citeyearpar{Berner2019} observe for Dota 2 (their comments also apply to Starcraft II): \textit{"it is infeasible for us to render each frame to pixels in all training games; this would multiply the computation resources required for the project many-fold"}. 


 


Several observations made in these papers are of interest. By using only behavioural cloning Vinyals et al.'s agent acheived a rank in the top 16\% of human players, showing behavioural cloning on a large enough scale can create strong agents. Wang et al. \citeyearpar{Wang2020a} conducted ablations of this work, finding skill-level of the replays was just as important as quantity. Kanervisto et al.  \citeyearpar{Kanervisto2020} found that combining data from different demonstrators can sometimes perform worse than training on only the strongest demonstrator's data.



\begin{table*}[b!]
\caption{Comparison of selected prior work using imitation learning in games.}
\label{tbl_prior_imitation}
\begin{center}
\resizebox{\textwidth}{!}{
\begin{tabular}{l l c l c l}
	\toprule
    Citation & Game & FPS? & Dataset size &  From pixels? & NN architecture \\
   \midrule 
   \citep{Harmer2018} & In-house game & \cmark & 45 minutes & \cmark & 4-layer CNN+LSTM \\
   \citep{Gorman2007} & Quake 2 & \cmark & 60 minutes & \xmark & 2-layer MLP\\
   \citep{Kanervisto2020} & Various incl. Doom & \cmark & 45 minutes & \cmark  & 2-layer CNN \\
   \hdashline
   \citep{Chen2017} & Super Mario Smash Bros & \xmark & 5 hours & \cmark  & 5-layer CNN+2-layer MLP\\
   \citep{Hester2018} & Atari & \xmark & 60 minutes & \cmark & 2-layer CNN+FC\\
   \citep{Bukaty2020} & NecroDancer & \xmark &  100 minutes & \cmark  & ResNet34 \\
   \citep{Vinyals2019} & Starcraft II & \xmark  & 4,000 hours & \xmark &  Mixed incl. ResNet, LSTM \\
   \citep{Silver2016} & Go & \xmark & 30 million frames  & \xmark &  Deep residual CNN \\
    \citep{Guss2019} & Minecraft & \xmark & 500 hours  & \xmark &  -- \\
   \hdashline
   Our work & CSGO & \cmark & 100 hrs, 5.8 million frames & \cmark  & EfficientNetB0+ConvLSTM\\
   \bottomrule
\end{tabular}
}
\end{center}
\end{table*}

\textbf{Counter-Strike specific work:} Despite its popularity, relatively little academic research effort has been applied to the Counter-Strike franchise, likely due to there being no API to conveniently interface with the game, and difficulty in mass roll-outs. Relevant machine learning works include predicting enemy player positions using Markov models \citep{Hladky2008}, and predicting the winner of match ups \citep{Makarov2018}. Other academic fields have studied the game and culture from various societal perspectives, e.g. \citep{Reer2014, Hardenstein2015}. Ours is the first academic work to build an AI from pixels for CSGO.


\textbf{Imitation learning challenges:} There has been an increasing awareness that leveraging existing datasets for tasks typically tackled through pure RL promises improved efficiency and will be valuable in many real-world situations -- a field labelled as offline RL \citep{Levine2020, Fu2020}, of which behavioural cloning is one approach.

Behavioural cloning systems can lead to several common issues, and recent research has aimed to address these -- e.g. agents may learn based on correlations rather than causal relationships \citep{Haan2019}, and accumulating errors can cause agents to stray from the input space for which expert data was collected \citep{Ross2011, Laskey2017}. One popular solution is to use a cloned policy as a starting point for other RL algorithms \citep{bakker1993, schaal1996}, with the hope that one benefits from the fast learning of behavioural cloning in the early stages, but without the performance ceiling or distribution mismatch. 

\newpage
\section{Interfacing with the Game}
\label{sec_app_interfacing}

A major challenge of the project was solving the engineering task of reliably interacting with the game. CSGO's code base is not open sourced and given a widespread cheating problem in CSGO, automated control has been restricted as far as possible.

\textbf{Image capture:} There is no direct access to CSGO's screen buffer, so the game must first be rendered on a machine and then pixel values copied. We used the Win32 API for this purpose -- other options tested were unable to operate at the required frame rate.

\textbf{Applying actions:}
Actions sent by many standard Python packages, such as pynput, were not recognised by CSGO. Instead, we used the Windows ctypes library to send key presses and mouse movement and clicks. 

\textbf{Recording local actions:} 
The Win32 API again was used to log local key presses and mouse clicks. Mouse movement is more complicated -- the game logs and resets mouse position at high-frequency and irregular intervals. Naively logging the mouse position at the required frame rate fails to reliably determine player orientation. We instead infer mouse movement from game metadata.

\textbf{Capturing game metadata:} 
CSGO provides a game state integration (\textcolor{blue}{\url{https://developer.valvesoftware.com/wiki/Counter-Strike:_Global_Offensive_Game_State_Integration}}) (GSI) API for developers, e.g. to enable automation of effects during live games. It is carefully designed to \textit{not} provide information that would give players an unfair advantage (such as location of enemy players). We use GSI to collect high-level information about the game, such as kills and deaths. Although it provides data about a player's state, we found this was not reliable enough to accurately infer mouse movements. 

For lack of alternatives, we parsed the local RAM to obtain precise information about a player's location, orientation and velocity. Whilst this approach is typically associated with hacking in CSGO, we only extract information about the player we are currently spectating, and never to provide the agent or its training data with information a human player would not have.


\newpage
\section{CSGO Game Settings}
\label{sec_app_running}

This section details the game settings we used to evaluate the agent. It's possible performance may drop if these are not matched, or if future CSGO updates materially affect gameplay -- we developed the agent over versions 1.37.7.0 to 1.38.0.1. 


\textbf{Note.} An update was released on 21 September 2021  (version 1.38.0.2) with two changes of relevance. 1) Several new variations of deathmatch can now be played (e.g. all vs. all, or team based). The version we consider in this paper is still available as `classic'. 2) There was a minor change to the `dust2' map, with a line of sight blocked from `T spawn'. Whilst this creates a small difference compared to the datasets collected in this paper, initial tests on the new modified version suggest this has a non-material impact on the agent's performance. A static version of the map used in this paper (and datasets) is available through the official steam workshop, \textcolor{blue}{\url{https://steamcommunity.com/sharedfiles/filedetails/?id=2606435621}}, which we'd recommend using for future benchmarking against the built-in bots.

\begin{list}{•}{}
\item CSGO version: 1.38.0.0
\item Game resolution: 1024$\times$768 (windowed mode)
\item Mouse sensitivity: 2.50
\item Mouse raw input: Off
\item Crosshair settings: Classic static, green -- RGB: (46, 250, 42), length 4.3, thickness 1.8, gap 2.0, no outline, no centre dot. (Length 2.8 also used in some training data and demos.)
\item Crosshair code: \verb|CSGO-UKcZG-QN8eW-WQMvd-NX6xr-RPqRP|
\item All graphics options: Lowest quality setting
\item Clear decals is bound to `n' key
\end{list}

\subsection{Game Modes Set-up}

The \textbf{aim train mode} uses the `Fast Aim / Reflex Training' map: \textcolor{blue}{\url{https://steamcommunity.com/sharedfiles/filedetails/?id=368026786}}, with the below console commands. Join counter-terrorist team, ensure that `God' mode is on, and AK47 is selected.
\begin{lstlisting}[breaklines=true]
sv_cheats 1; 
sv_pausable 1; 
mp_roundtime 6000; 
sv_infinite_ammo 1;
fps_max 64;
sv_auto_adjust_bot_difficulty 0;
\end{lstlisting}

 \textbf{Easy and medium deathmatch mode} can be initialised from the main menu (deathmatch$\to$ play offline with bots $\to$ dust\_2 $\to$ easy/medium bots $\to$ classic mode). Then run below commands (have to be run in several batches). Manually join terrorist team, and select AK47. Note we observed some variability between the difficulty levels even under this procedure -- to improve repeatability of our reported results we exited/restarted CSGO at least three times. 

\textbf{Deathmatch mode, easy setting}
\vspace{-0.05in}
\begin{lstlisting}[breaklines=true]
sv_cheats 1; 
sv_pausable 1; 
mp_roundtime 6000;
sv_infinite_ammo 1;
mp_teammates_are_enemies 0;
mp_limitteams 30;
mp_autoteambalance 0;
fps_max 64;
bot_kick;

bot_pistols_only 1;
bot_difficulty 0;
sv_auto_adjust_bot_difficulty 0;
contributionscore_assist 0;
contributionscore_kill 0;
mp_restartgame 1;

bot_add_t; (run 11 times)
bot_add_ct; (run 12 times)
\end{lstlisting}

\textbf{Deathmatch mode, medium setting}
\vspace{-0.05in}
\begin{lstlisting}[breaklines=true]
sv_cheats 1; 
sv_pausable 1; 
mp_roundtime 6000;
sv_infinite_ammo 2;
mp_teammates_are_enemies 0;
mp_limitteams 30;
mp_autoteambalance 0;
fps_max 64;
bot_kick;

bot_difficulty 1;
sv_auto_adjust_bot_difficulty 0;
contributionscore_assist 0;
contributionscore_kill 0;
mp_restartgame 1;

bot_add_t; (run 11 times)
bot_add_ct; (run 12 times)
\end{lstlisting}

\textbf{Deathmatch mode, human setting} is initialised from the main menu (play online deathmatch $\to$ dust\_2 $\to$ classic mode). Manually join terrorist team, and select AK47. Note that there does seem to be some form of skill-based and trust-based matchmaking for deathmatch mode, so the specific skill-level of opponents will vary for different accounts. We use an account with `prime status', but presume it has a low skill-level since we use the same account to scrape the online dataset.

\newpage
\section{Miscellaneous}
\label{sec_app_misc}

\subsection{CSGO's Full Action Space}

\begin{table}[h]
\begin{center}
\resizebox{0.6 \textwidth}{!}{
\begin{tabular}{l l c l}
\toprule
    Action &  Meaning & Output by agent? & Output activation  \\
   \midrule 
    w,a,s,d & forward, backward, left, right & \cmark & sigmoid \\
    space   & jump & \cmark & sigmoid \\
    r       & reload & \cmark &  sigmoid \\
    e    & use (e.g. open door) & \xmark & -- \\
    ctrl    & crouch & \xmark & -- \\
    shift   & walk & \xmark & -- \\
    1,2,3,4   & weapon switch & \xmark & -- \\
    left click   & fire & \cmark & sigmoid \\
    right click   & zoom & \xmark & -- \\
    mouse x \& y & aim & \cmark &  2$\times$softmax \\
  \hdashline
    value & value estimate & \cmark &  linear \\
    \bottomrule
\end{tabular}
}
\end{center}
\caption{CSGO action space, compared with the agent's output space.}
\label{tbl_action_space}
\end{table}

\end{document}